\documentclass[pmlr]{jmlr}

\usepackage{longtable}%

\usepackage{booktabs}

\usepackage{CJKutf8}
\usepackage{makecell}
\usepackage{booktabs}
\usepackage{multirow}
\usepackage{xpinyin}
\usepackage{graphicx}
\usepackage{tipa}
\usepackage{color}

\usepackage{amsmath}
\usepackage{graphicx}
\makeatletter
\protected\def\my@emoji@pic #1#2{\leavevmode@ifvmode
\lower\dimexpr #1\p@*1/10\hbox{\includegraphics[height={#1\p@}]{#2}}}
\def\my@emoji@math #1{%
\mathchoice
{\my@emoji@pic\tf@size{#1}}{\my@emoji@pic\tf@size{#1}}
{\my@emoji@pic\sf@size{#1}}{\my@emoji@pic\ssf@size{#1}}}
\protected\def\myemoji #1{{\ifmmode\my@emoji@math{#1}\else\my@emoji@pic\f@size{#1}\fi}}
\makeatother
\newcommand{\screw}{\myemoji{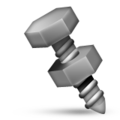}}

\ExplSyntaxOn
\NewDocumentCommand{\splitautopy}{}
  {
    \group_begin:
      \tl_gset_eq:Nc \l_pinyin_tl
        {
           c__xpinyin_ \int_to_arabic:n { `好 } _tl
        } 
      \__split_pinyin:V \l_pinyin_tl         
    \group_end:
  }
  
\NewDocumentCommand{\splitmanulpy}{}
  {
    \group_begin:
      \tl_gset:Nx \l_pinyin_tl
        {
           \pinyin{hao3}
        } 
      \__split_pinyin:V \l_pinyin_tl
    \group_end:
  }  

\tl_new:N \l_pinyin_tl
\tl_new:N \l_test_pinyin_tl
\str_new:N \l_pinyin_str

\cs_new_protected:Npn \__split_pinyin:n #1
  {
    \str_clear:N \l_pinyin_str
    \tl_clear:N  \l_test_pinyin_tl
    \tl_set:Nn \l_test_pinyin_tl {#1}

    \tl_map_inline:Nn \l_test_pinyin_tl
      {
        \str_put_right:Nn \l_pinyin_str {##1}
      }
    \str_use:N \l_pinyin_str
  }
\cs_generate_variant:Nn  \__split_pinyin:n { V }
\cs_generate_variant:Nn  \__split_pinyin:n { x }
\ExplSyntaxOff

\usepackage{lineno}

\newcommand{\cs}[1]{\texttt{\char`\\#1}}
\makeatletter
\let\Ginclude@graphics\@org@Ginclude@graphics 
\makeatother

\jmlrvolume{260}
\jmlryear{2024}
\jmlrworkshop{ACML 2024}
\jmlrpages{-}
\renewcommand{\thepage}{}

\title[Chain Association-based Attacking and Shielding Natural Language Processing Systems]{Chain Association-based Attacking and Shielding Natural Language Processing Systems}

 \author{\Name{Jiacheng Huang} \Email{D190201005@stu.cqupt.edu.cn}\\
  \Name{Long Chen \nametag{\thanks{corresponding author}}} \Email{chenlong@cqupt.edu.cn}\\
  \addr School of Computer Science and Technology, Chongqing University of Posts and Telecommunications, 400065, Chongqing, China}

\editors{Vu Nguyen and Hsuan-Tien Lin}

\begin{document}

\maketitle

\begin{abstract}
Association as a gift enables people do not have to mention something in completely straightforward words and allows others to understand what they intend to refer to.
In this paper, we propose a chain association-based adversarial attack against natural language processing systems, utilizing the comprehension gap between humans and machines.
We first generate a chain association graph for Chinese characters based on the association paradigm for building search space of potential adversarial examples. 
Then, we introduce an discrete particle swarm optimization algorithm to search for the optimal adversarial examples. 
We conduct comprehensive experiments and show that advanced natural language processing models and applications, including large language models, are vulnerable to our attack, while humans appear good at understanding the perturbed text.
We also explore two methods, including adversarial training and associative graph-based recovery, to shield systems from chain association-based attack.
Since a few examples that use some derogatory terms, this paper contains materials that may be offensive or upsetting to some people.

\end{abstract}
\begin{keywords}
adversarial examples; chain associations; natural language processing; black box
\end{keywords}

\section{Introduction}
In past years, many studies has shown that the adversarial examples can cause decision-making errors in natural language processing (NLP) systems~\cite{formento-etal-2023-using,DBLP:journals/kais/OuYTC22}, even in large language models~\cite{DBLP:journals/debu/0001HH0ZWY0HGJ024} (LLMs).

Howerver, existing adversarial attacks only consider the attack strategies in a direct way while ignoring the complexity of textual adversarial attacks in reality.
For example, Chinese words \begin{CJK}{UTF8}{gbsn}``幼稚''\end{CJK}, which means ``\emph{naive}'', is an adjective with emotional tendency, but it will not be recognized as an emotional word by an emotion analysis system after being substituted with \begin{CJK}{UTF8}{gbsn}``拿衣服''\end{CJK}, which is a verb object phrase meaning to ``\emph{take clothes}''.
The relation between \begin{CJK}{UTF8}{gbsn}``幼稚''\end{CJK} and \begin{CJK}{UTF8}{gbsn}``拿衣服''\end{CJK} is not a simple single-layer mapping, but a multi-layer mapping. 
Specifically, we associate \begin{CJK}{UTF8}{gbsn}``幼稚''\end{CJK} with ``\emph{naive}'' first, which is the English translation of \begin{CJK}{UTF8}{gbsn}``幼稚''\end{CJK}, and then further associate it with \begin{CJK}{UTF8}{gbsn}``拿衣服''\end{CJK}, which is one of the Mandarin transliterations of ``\emph{naive}''.
Note that the above is only a simple example of word substitution based on chain association while the associative ability of human being is complex.

English has analogous cases. Take ``screw'', a polysemy referring to ``a metal object like a nail'' or ``having sex with someone''. 
Attackers online can replace ``screw'' with its related emoji to form offensive phrases like ``$\screw$ you'', which utilizes human associations about the corresponding word of emoji and its polysemy.
Note that this example is merely used as an analogy to explain our idea, in fact, this work only focuses on Chinese adversarial examples.

In this work, we investigate to what extent advanced Chinese NLP systems are sensitive to chain association-based attack and explore various shielding techniques.
Specifically, we first generate a chain association graph for Chinese characters based on the association paradigm for building search space of potential adversarial examples. 
Then, we regard generating adversarial examples as a problem of combinatorial optimization and introduce an discrete particle swarm optimization algorithm to search for the optimal adversarial examples. 
We show that advanced NLP models and applications are extremely vulnerable to our attack. 
To our best knowledge, we are the first to introduce chain association in adversarial attack. 
Furthermore, we also explore two methods to protect NLP systems from our attacks. 

 \section{Related Work}
 Although adversarial attacks were first proposed in the field of image recognition,
 however, due to the discrete nature of textual data and the uncertainty brought by perturbations to the semantic quality of text, it is difficult to directly migrate adversarial attack algorithms in the image domain to textual data types.
 Therefore, researchers have conducted extensive studies on textual data, forming different textual adversarial attack paradigms with different perturbation granularities, such as character-level, word-level, and sentence-level.
 The form of character-level attacks vary across different linguistic and cultural environments. 
 In the English context, character-level attacks often exploit visual perturbations, including artificially constructed spelling errors such as the insertion, deletion, swapping, and modification of letters within words~\cite{formento-etal-2023-using}. 
 However, in the Chinese context, handwritten stroke errors on paper do not occur in electronic devices based on input methods. 
 Therefore, character-level attacks in Chinese environments usually manifest as the replacement of homophones~\cite{DBLP:journals/access/ChengCGPZ20} or the utilization of visual character disassembly~\cite{DBLP:journals/kais/OuYTC22}.
  
 Word-level adversarial attacks achieve the deviation of semantic vectors of examples by disturbing the input at the word level, making them cross the decision boundary and thus leading to incorrect model outputs. 
 As the core method of this strategy, word substitution covers various strategies, including word vector similarity~\cite{DBLP:conf/aaai/JinJZS20}, synonyms~\cite{RenDHC19}, sememes~\cite{zang2020}, and language model scores~\cite{zhang19}.
 Word-level adversarial attacks do not violate the grammatical rules of the text and maximize the retention of the original semantics, thus exhibiting better performance in terms of adversarial text quality and attack success rate. Additionally, the utilization of language models for control also ensures the fluency and smoothness of the adversarial text.
  
 Sentence-level adversarial attacks treat the entire original input sentence as the target of perturbation, carefully reconstructing the textual content by generating adversarial texts that maintain the same semantic meaning as the original input but cause the victim model to make incorrect decisions.
 Common sentence-level adversarial attack methods include re-encoding and decoding after encoding~\cite{DBLP:conf/emnlp/HanZJT20}, adding irrelevant sentences~\cite{DBLP:conf/ijcai/0002LSBLS18}, and paraphrasing~\cite{xu-etal-2021-grey}.
 Sentence-level adversarial attacks face greater difficulties in maintaining the original semantics.
  
 Our work is a kind of Chinese multi-level perturbation based on the gift of human beings namely chain associations, which is utilizing the comprehension gap between humans and machines in cognition.
  
 \section{Associations in Textual Adversarial Examples}
 \subsection{Motivation}
 Abundant association ability is vested in human beings, who are able to associate things that seem totally different but related.
 The gift enables people do not have to mention something in completely straightforward words and allows others to understand what they intend to refer to.
 Associationism psychology holds that all complex mental processes, such as thinking, learning, and memory, can be mainly explained by the associative links that connect ideas, according to specific laws and principles~\cite{associationism}, 
 e.g., Philosopher David Hume's Laws of Association: (i) Law of Similarity, (ii) Law of Contiguity, and (iii) Law of Causality.
 The Law of Similarity states that when two things are very similar to each other, the thought of one will often trigger the thought of the other.
 The Law of Contiguity states that we associate things that occur close to each other in time or space.
 The Law of Causality states that we associate things when there is a causal relationship with them.
 
 We believe that the distance between associative words is close in human cognition even if it is far in word meaning, 
 and such an inconsistency between two kinds of distances of associative words provides the motivation for this paper.
 Specifically, the existing textual adversarial attacking strategies, such as word substitutions and misspellings, are special cases of the laws of association.
 For examples, the synonym-based substitution belongs to the law of similarity in word meaning and the misspellings belongs to the law of similarity in vision.
 All these strategies take advantage of the inconsistency between the distances in human cognition and word meaning.
 
 Furthermore, the association of words is not necessarily single-layer as a chain association chain will be formed while associating constantly from one word to another.
 We believe that such a chain association of words will aggravate the inconsistency of distances in human cognition and word meaning,
 and cause the deep neural networks to fail blatantly.
 
 \subsection{Rules of Word Association}
 It should be noted that not all associations can be used in the field of textual adversarial examples, as inappropriate associative word substitutions will cause the text to be unreadable.
 Thus, we summarize several rules of word association can be used in textual adversarial attack, according to Philosopher David Hume's laws of association and Chinese cultural environment, as shown in Table~\ref{table:rules}.
 The Law of Contiguity is excluded because it is difficult to introduce this law into textual adversarial attack.
 
 \begin{table}[htbp]
   \centering
   \caption{\label{table:rules} Rules of Word Association based on David Hume's Laws of Association. 
   The detailed explanations and implementation methods can be see in section~\ref{sec:graph}.}
   \begin{tabular}{cc}
     \hline
     \textbf{Laws of Association}                & \textbf{Rules of Word Association}               \\ \hline
     \multirow{5}{*}{Law of Similarity} & English Translation         \\ \cline{2-2} 
                                 & Transliteration             \\ \cline{2-2} 
                                 & Fuzzy Matching              \\ \cline{2-2} 
                                 & Visually Similar Characters \\ \cline{2-2} 
                                 & Characters Disassembling    \\ \hline
     \multirow{3}{*}{Law of Causality}  & Chinese Pinyin              \\ \cline{2-2} 
                                 & Acronym                     \\ \cline{2-2} 
                                 & Hanzify                     \\ \hline
     \end{tabular}
   \end{table}
 
 With the help of knowledge graph technology, the association chain can be fully expressed while entity refers to word and edge refer to the associative relationship between words.
 Words out of vocabulary may appear while associating words in some way, such as the example mentioned in the introduction.
 Although these words have no specific meaning, we believe that readers will start a word guessing process which is kind of like completing a cloze, i.e., mask the unknown words temporarily and infer what the author intends to write after reading context.
 Due to the association between the original word and guessed word, readers can confirm whether they guessed correctly.
 Moreover, the word guessing process will be easier than normal cloze, since the topics involving supervision are usually confined to crime, pornography, and dirty words.
 
 \section{Approach}
 Generally, words in the original sentence vary in their impact on model predictions. Minor sentence alterations, especially replacing key words, can significantly alter predictions.
 Following \cite{DBLP:conf/ndss/LiJDLW19}, we identify the most influential words by measuring their removal effects and replace them in order of importance.
 The challenge lies in generating suitable substitutes and determining optimal adversarial examples that deceive the model while resembling the original.
 Here, we propose 1) an associative knowledge graph and 2) an adversarial search strategy.
 
 \subsection{Associative Knowledge Graph}
 \label{sec:graph}
\begin{figure}[htbp]
    \begin{center}
    \includegraphics[width=0.9\textwidth]{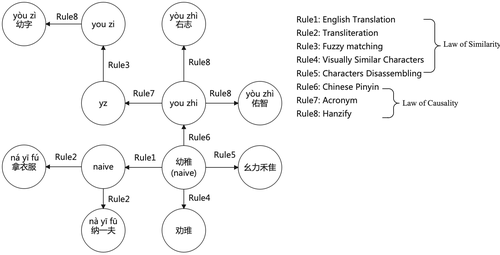}
    \caption{\label{fig:rules} An example of chain-association based knowledge graph.}
    \end{center}
    \end{figure}

 The associations can be represented by a graph intuitively so that we consider building an associative knowledge graph $\mathit{G}$ as shown in Fig.~\ref{fig:rules}. 
 In fact, due to the complexity of human association ability, we can not completely enumerate all the types of word association but summarize some typical rules to test our ideas. 
 Fortunately, our design is highly scalable, allowing new rules and graph updates. 
 Next, we elaborate on these rules and their implementation.
 
 \noindent\textbf{English Translation.} \ We view English translations of Chinese characters as word associations based on similarity of meaning. For example, ``\emph{naive}'' translates to \begin{CJK}{UTF8}{gbsn}``幼稚''\end{CJK}. 
 Many Chinese online users know both languages, and English is crucial in Chinese education. English translations can be accessed via third-party platforms.
 
 \noindent\textbf{Chinese Pinyin.} \ Chinese Pinyin is the Chinese phonetic notation with Latin alphabets according to Laws of Causality while each Chinese character has its corresponding Chinese Pinyin.
 For instance, it can be written as ``\emph{you zhi}'' if we were to represent the pronunciation of the Chinese characters \begin{CJK}{UTF8}{gbsn}``幼稚''\end{CJK} in Chinese Pinyin.
 The conversion between Chinese characters and Chinese Pinyin can be implemented with the Python third-party library named pypinyin.
 
 \noindent\textbf{Transliteration.} \ Transliteration refers to the translation of foreign words with Chinese characters with similar pronunciations and it is based on Laws of Similarity.
 This kind of Chinese characters used for transliteration no longer have their original meaning but only retain their pronunciation and writing form.
 For example, consider the pronunciation of the word ``\emph{naive}'' as /\textipa{naI"i:v}/. 
 This pronunciation is akin to the Chinese characters \begin{CJK}{UTF8}{gbsn}``拿(ná)衣(yī)服(fú)''\end{CJK} in terms of phonetics. 
 To be more precise, /\textipa{naI}/ is similar to ``\emph{ná}'', /\textipa{i}/ is reminiscent of ``\emph{yī}'', and /\textipa{v}/ bears resemblance to ``\emph{fú}''.
 We only consider the transliteration from English to Chinese in this work and implement it by establishing the mapping relationship between English phonetics and Chinese Pinyin, as the number of vowels and consonants in English is limited and their pronunciations can be similarly expressed with Chinese pinyin.
 
 \noindent\textbf{Acronym.} \ Acronym namely a word composed of the first letters of the words in a phrase, which is often used as an abbreviated form of Chinese network language, can be used as an association word of Chinese pinyin or English word.
 It is based on Law of Causality as the acronym cause readers want to figure out what the complete phrase is.
 For example, ``lol'' is typically an abbreviation for ``laugh out loud'' in English.
 Similarly, in Chinese internet slang, ``nb'' is often an abbreviation for \begin{CJK}{UTF8}{gbsn}``牛(niú)逼(bī)''\end{CJK}, which means awesome.
 The conversion from Chinese Pinyin or English word to its acronym can be implemented with built-in Python string operations.
 
 \noindent\textbf{Fuzzy Matching.} \ Fuzzy matching for Chinese Pinyin, i.e., search possible Pinyin starting with given first letters (i.e., acronym), is a common substitution for keywords in Chinese offensive text online recently, which often escapes offensive text automated detection systems.
 For instance, due to their shared first letters in Pinyin, \begin{CJK}{UTF8}{gbsn}``特(tè)喵(miāo)的(de)''\end{CJK} in Chinese internet slang carries the same meaning as \begin{CJK}{UTF8}{gbsn}``他(tā)妈(mā)的(de)''\end{CJK} which is an offensive term equivalent to ``fuck'', while \begin{CJK}{UTF8}{gbsn}``特(tè)喵(miāo)的(de)''\end{CJK} originally lacks a specific meaning of its own.
 It is based on Law of Similarity because they have similar consonants.
 The operation can be implemented by arranging and combining the first letters and vowels since the first letters are often consonants.
 
 \noindent\textbf{Hanzify.} \ Hanzify, namely conversion from Chinese pinyin to Chinese characters in this paper, can also be an association link connected behind the fuzzy matching in the association chain, which is implemented simply with Python third party library named Pinyin2Hanzi.
 For example, the Pinyin ``hao'' can be converted into various Chinese characters, such as \begin{CJK}{UTF8}{gbsn}``好''(good)\end{CJK}, \begin{CJK}{UTF8}{gbsn}``号''(number)\end{CJK}, and \begin{CJK}{UTF8}{gbsn}``郝''(a Chinese surname)\end{CJK}.
 It is based on Law of Causality as Chinese Pinyin cause readers to associate the corresponding Chinese characters.
 
 \noindent\textbf{Visually Similar Characters.} \ Visual similarity can play a considerable role in replacing the important words in a sentence according to Law of Similarity, as a large number of characters with visual similar composition or shape exist in Chinese.
 For instance, the shape of the character \begin{CJK}{UTF8}{gbsn}``日''\end{CJK} is similar to the character \begin{CJK}{UTF8}{gbsn}``曰''\end{CJK} visually, even though they have completely different meanings. 
 Specifically, the character \begin{CJK}{UTF8}{gbsn}``日''\end{CJK} means ``day'' in a general context or ``fuck'' in an offensive context, while \begin{CJK}{UTF8}{gbsn}``曰''\end{CJK}  means ``say'' in classical Chinese language.
To this end, about twenty thousand Chinese characters are collected and converted into image representation using Python built-in library named PIL, and we build character embedding space with visual shape for retrieving similar characters. 
 
 \noindent\textbf{Characters Disassembling.} \ Chinese characters include a large number of combined characters composed of several single characters, some of which can be disassembled transversely and not or only mildly affect reading as the characters are left-right structures according to Law of Similarity.
 For example, Chinese character \begin{CJK}{UTF8}{gbsn}``幼''\end{CJK} can be disassembled into \begin{CJK}{UTF8}{gbsn}``幺''\end{CJK} and \begin{CJK}{UTF8}{gbsn}``力''\end{CJK}, i.e., \begin{CJK}{UTF8}{gbsn}``幺力''\end{CJK}.
 Similarly, the Chinese character \begin{CJK}{UTF8}{gbsn}``稚''\end{CJK} can be disassembled into \begin{CJK}{UTF8}{gbsn}``禾''\end{CJK} and \begin{CJK}{UTF8}{gbsn}``隹''\end{CJK}, i.e., \begin{CJK}{UTF8}{gbsn}``禾隹''\end{CJK}.
 We collected the required data by crawler from a third-party website providing characters disassembling service.
 
 Using the above rules, we can generate a large number of candidates for any word that needs to be replaced in a sentence.
 In order to modify the original sentence as little as possible, we only take the words whose importance is the one-third of the original sentence as the substituted words in this work.
 
 \subsection{Particle Swarm Optimization-based Search Strategy}
 The original sentences may contain multiple important words to be replaced and, for each word, abundant associative words exist according to the association rules.
 Besides, as we can see in Fig.~\ref{fig:rules}, there are multiple paths and multiple layers to retrieve the associative words of a original word. 
 Thus, an efficient adversarial examples search strategy is essential while a huge search space composed of numerous potential adversarial examples needs to be handled and, furthermore, the path from the original word to the final associative word is expected to be as short as possible for reducing the burden of comprehension to readers.
 Hence, we consider exploiting particle swarm optimization (PSO) to search for final adversarial examples in the search space.
 
 However, rather than the original PSO, whose search space is continuous, a variation suitable for discrete space composed of potential textual adversarial examples is what we need.
 Inspired by~\cite{zang2020}, a \emph{position} in the search space now corresponds to a potential adversarial example while each dimension of a \emph{position}, i.e., sentence, corresponds to a word.
 Besides, the \emph{velocity} of particle now corresponds to a probability vector, and each dimension of a velocity refers to the corresponding dimension of the related \emph{position} will change, i.e., the probability of a word will be substituted.
 
 Formally, we denote an original sentence as $x^{o}=x^{o}_{1}...x^{o}_{n}...x^{o}_{N}$, where $N$ is the length of original sentence and $x^{o}_{n}$ is the $n$-th word in original sentence.
 A position in the search space, i.e., a potential textual adversarial example related to $x^{o}$, is denoted as $x^{a}=x^{a}_{1}...x^{a}_{n}...x^{a}_{N}, x^{a}_{n} \in \mathbb{A}(x^{o}_{n})$, where $\mathbb{A}(x^{o}_{n})$ is the set of $x^{o}_{n}$ and all its associative words.
 A velocity of a particle, is denoted as $v={v^{}_{1}...v^{}_{n}...v^{}_{N}}$, where $v^{}_{n}$ refers to the probability with which determines whether $n$-th dimension of the particle's position will move.
 
 Since we expect that the final adversarial example can not only fool the victim model but also the association paths between substitution words and original words are as short as possible in total, the optimization score of a position is calculated by following the formula:
 
 \begin{equation}
 score(x^{a})= \frac{1-C(x^{a})}{L(x^{o}, x^{a})}
 \end{equation}
 
 where $C(x^{a})$ is the confidence of the true label of $x^{a}$ given by the victim model, and $L(x^{o},x^{a})$ is the total number of layers from the associative words in $x^{a}$ used to replace original words in $x^{o}$.
 If the current number of iterations reaches the maximum, the algorithm will terminate and output the position of the particle in the \emph{global best previous position} as the search result.
 
 When presented with an original sentence, each particle is given a stochastic \emph{position} $x$ and \emph{velocity} $v$. 
 Specifically, the important words in the original sentence are substituted with their direct associative words and take the modified sentence as the initial \emph{position} of a particle.
 In addition, the \emph{velocity} $v$ of particle in discrete search space is a probability vector, hence, each dimension of $v$ is initialized randomly between interval $[0.0,1.0]$ and updated by the following formula:
 
 \begin{equation}
   \begin{aligned}
 &v_{n}=S(\omega v_{n}+\varphi^{}_{1} I(p,x)+\varphi^{}_{2} I(p^{g},x))\\
 &I(a,b)=  \left\{ 
   \begin{array}{l}
    -1,\ \ a = b \\
    \ \ \ 1, \ \ a \neq b \\
   \end{array}
   \right.
   \\
 \end{aligned}
 \end{equation}
   
 where $S(*)$ is a sigmoid limiting transformation for constraining $v_{n}$ to the interval $[0.0,1.0]$ since $v_{n}$ is a probability.
 It can be reasonably calculated as that $v_{n}$ is going to decrease, remain the same, or increase when the particle's position is at the global best previous position, individuals best previous position only, or both not.
   
 In addition, compared with the fixed value, a dynamic decreasing $\omega$ derived from a measure function enables particles to explore more positions in the early stage and gather around the best positions in the final stage.
 Thus, we introduce a nonlinear dynamic adaptation function to update $\omega$.
 Specifically, 
   
 \begin{equation}
    \omega = \omega_{max}-\frac{t^{2}\times(\omega_{max}-\omega_{min})}{t^2_{max}}
 \end{equation}
 
 where $t$ and $t_{max}$ are the current and max numbers of iterations separately.
 $\omega$ decreases slowly during the initial iteration, which is conducive to exploring the local optimum at an early stage, and $\omega$ decreases rapidly while approaching the maximum number of iterations for improving the efficiency of converging to the global optimum.
 
 Besides, the original position update formula that makes addition is also not suitable for discrete space.
 Inspired by~\cite{kennedy1997discrete}, we propose a probabilistic approach to update the position of particles.
 A probability $P$ is introduced with which a particle determines whether one of its position's dimensions, i.e., $x_{n}$, moves to the corresponding dimension of \emph{global best previous position} $p_{n}^{g}$.
 The movement of each dimension of a particle' position at an iteration is redefined by the following rule:
 
 $$
 \begin{aligned}
 &if\ rand()<P\ and\ rand()<v_{n}\\
 &then\ x_{n}=p_{n}^{g}\\
 &else\ rand()>P\ and\ rand()<v_{n}\\
 &then\ x_{n}= G_{adj}(x_{n})
 \end{aligned}
 $$
 
 where $rand()$ refers to a random number selected from a uniform distribution in $[0.0,1.0]$, and $G_{adj}(x_{n})$ refers to one of the words adjacent to $x_{n}$ at random in the associative graph $\mathit{G}$.
 In addition, to encourage particles to explore more positions according to associative graph $G$ at an early stage and search for better positions around the global best position at a final stage, $P$ varies with iteration as follow:
 
 \begin{equation}
   P=P_{min}+\frac{t^2\times (P_{max}-P_{min})}{t^2_{max}}
 \end{equation}
 
 where $0<P_{min}<P_{max}<1$, and we can see the probability of particles moving to the \emph{global previous best position} will increase with the number of iterations.

 \section{Experiments}
 In this section, we conduct comprehensive experiments to evaluate our chain association-based attack on Chinese NLP systems and study the methods of shielding against chain association-based adversarial attack.
 \subsection{Attacking}
 \noindent \textbf{Victims Models and Applications.} \ To investigate the effects of chain association-based attack, we evaluate our attack method on five text classification models, namely Fasttext~\cite{Fasttext}, TextCNN~\cite{DBLP:journals/ijon/GuoZLM19}, Attention-based Bidirectional LSTM (BiLSTM)~\cite{DBLP:journals/access/XiaoyanR23} and BERT~\cite{DBLP:journals/jcsc/SiW23}. 
 Besides, we also test our attack methods on two industry-leading commercial applications used for offensive text detection, namely Baidu Text Censoring\footnote{https://ai.baidu.com/solution/censoring} (BaiduCensor) and Alibaba Content Security\footnote{https://homenew.console.aliyun.com} (AliSecurity).
 
 In addition, due to the significant impact of large-scale models on artificial intelligence recently, we also conduct experiments on LLMs, i.e., ERNIE Bot and Tongyi Qianwen, released by Baidu and Alibaba respectively.
 Due to the outputs of some LLMs are not easy to control while using the prompt-based paradigm, it is difficult to apply our attack algorithm directly to LLMs while each iteration in our adversarial attacking algorithm requires getting the confidence of classification of the victim model. 
 Thus, as an alternative, we used the paradigm of transfer-based attack to conduct experiments on LLMs, that is, we obtain adversarial examples generated on the local model, and then use prompt-based paradigm to get answers for classification of LLMs by inputting prompt.
 All prompts used in this paper are presented in Appendix~\ref{sec:prompts}.
 
 \noindent \textbf{Dataset and Associative Graph.} \ We use public Chinese datasets, i.e., Meituan~\footnote{Meituan is a platform for ordering takeaway, which contains positive and negative user comments} and Amazon comments.
 Besides, we collect offensive text and normal text from online social platforms labeled by native Chinese speakers. 
 We divide the dataset into two parts, i.e., 80\% for training and 20\% for testing.
 Details of the datasets are shown in Table~\ref{table:dataset}.

  \begin{table}[htbp]
     \centering
     \caption{\label{table:dataset} Statistics for the datasets. `Avg.Len' refer to the average length of examples.}
     \begin{tabular}{ccccc}
     \toprule
     \makecell[c]{\textbf{Dataset}}&\makecell[c]{\textbf{Class}}&\makecell[c]{\textbf{Avg.Len}}&\makecell[c]{\textbf{Train}}&\makecell[c]{\textbf{Test}}\\
     \midrule
     \makecell[c]{Meituan Comments}&\makecell[c]{2}&\makecell[c]{19}&\makecell[c]{9600}&\makecell[c]{2400}\\
     \makecell[c]{Amazon Comments}&\makecell[c]{2}&\makecell[c]{24}&\makecell[c]{9600}&\makecell[c]{2400}\\
     \makecell[c]{Offensive Text}&\makecell[c]{2}&\makecell[c]{51}&\makecell[c]{8000}&\makecell[c]{2000}\\
     \bottomrule
     \end{tabular}
     \end{table}
 
 \noindent \textbf{Baseline Methods.} \ To evaluate our chain association-based adversarial attack more comprehensively, we implemented two baseline methods and compared them with ours.
 The two baseline methods are 1) GreedyAttack~\cite{DBLP:journals/kais/OuYTC22}, and 2) WordChange~\cite{DBLP:journals/access/ChengCGPZ20}, both of which represent multi-strategy approaches for generating Chinese adversarial examples.

 \noindent \textbf{Attacking Performance.} \ The performances and attack results of all models and applications on Meituan, Amazon and offensive text are listed in Table~\ref{table:performance_mt}, Table~\ref{table:performance_am} and Table~\ref{table:performance} respectively.
 ``ACC'' refers to the accuracy of models and applications in different conditions and ``WMD'' refers to Word Mover's Distance between original text and perturbed text.
 We observe that our chain association-based adversarial attacking method decreases the accuracy of the victim models and applications the most compared with the other two baseline methods.
 For example, it attacks Baidu Text Censoring's service and reduce its accuracy from 96.4\% to 25.2\% notably for offensive text detection, which demonstrates the vulnerability of existing offensive text automated detection applications.
 
 Besides, the attack results of LLMs are shown in Table~\ref{table:performance_llm}, which demonstrates that LLMs can be also affected by the adversarial examples generated by our method.
 But we also observe that the performances of adversarial attacks on LLMs when applied to offensive text are not significant.
 This could be due to the reason that the offensive texts contain a high degree of offensive context, which enables LLMs to maintain their robustness in the face of such kinds of adversarial examples.
 
 \begin{table}[htbp]
   \centering
   \caption{\label{table:performance_mt} Attack performances of different attack methods against victim models on Meituan Comments.
   }
 \begin{tabular}{ccccccccc}
   \toprule
   \multirow{2}{*}{\textbf{Model}}&\multicolumn{2}{c}{\textbf{Original}}&\multicolumn{2}{c}{\textbf{GreedyAttack}}&\multicolumn{2}{c}{\textbf{WordChange}}&\multicolumn{2}{c}{\textbf{Ours}}\\
   \makecell[c]{}&\makecell[c]{ACC}&\makecell[c]{WMD}&\makecell[c]{ACC}&\makecell[c]{WMD}&\makecell[c]{ACC}&\makecell[c]{WMD}&\makecell[c]{ACC}&\makecell[c]{WMD}\\
   \midrule
   \makecell[c]{Fasttext}&\makecell[c]{0.858}&\makecell[c]{N/A}&\makecell[c]{0.350}&\makecell[c]{0.536}&\makecell[c]{0.364}&\makecell[c]{0.528}&\makecell[c]{\textbf{0.184}}&\makecell[c]{0.524}\\
   \makecell[c]{TextCNN}&\makecell[c]{0.885}&\makecell[c]{N/A}&\makecell[c]{0.396}&\makecell[c]{0.494}&\makecell[c]{0.375}&\makecell[c]{0.497}&\makecell[c]{\textbf{0.227}}&\makecell[c]{0.499}\\
   \makecell[c]{BiLSTM}&\makecell[c]{0.894}&\makecell[c]{N/A}&\makecell[c]{0.374}&\makecell[c]{0.521}&\makecell[c]{0.371}&\makecell[c]{0.485}&\makecell[c]{\textbf{0.279}}&\makecell[c]{0.492}\\
   \makecell[c]{BERT}&\makecell[c]{0.934}&\makecell[c]{N/A}&\makecell[c]{0.543}&\makecell[c]{0.498}&\makecell[c]{0.528}&\makecell[c]{0.534}&\makecell[c]{\textbf{0.328}}&\makecell[c]{0.503}\\
   \bottomrule
   \end{tabular}
   \end{table}
 
 \begin{table}[htbp]
   \centering
   \caption{\label{table:performance_am} Attack performances of different attack methods against victim models on Amazon Comments.
   }
 \begin{tabular}{ccccccccc}
   \toprule
   \multirow{2}{*}{\textbf{Model}}&\multicolumn{2}{c}{\textbf{Original}}&\multicolumn{2}{c}{\textbf{GreedyAttack}}&\multicolumn{2}{c}{\textbf{WordChange}}&\multicolumn{2}{c}{\textbf{Ours}}\\
   \makecell[c]{}&\makecell[c]{ACC}&\makecell[c]{WMD}&\makecell[c]{ACC}&\makecell[c]{WMD}&\makecell[c]{ACC}&\makecell[c]{WMD}&\makecell[c]{ACC}&\makecell[c]{WMD}\\
   \midrule
   \makecell[c]{Fasttext}&\makecell[c]{0.898}&\makecell[c]{N/A}&\makecell[c]{0.346}&\makecell[c]{0.476}&\makecell[c]{0.453}&\makecell[c]{0.482}&\makecell[c]{\textbf{0.327}}&\makecell[c]{0.498}\\
   \makecell[c]{TextCNN}&\makecell[c]{0.900}&\makecell[c]{N/A}&\makecell[c]{0.375}&\makecell[c]{0.490}&\makecell[c]{0.490}&\makecell[c]{0.480}&\makecell[c]{\textbf{0.335}}&\makecell[c]{0.512}\\
   \makecell[c]{BiLSTM}&\makecell[c]{0.936}&\makecell[c]{N/A}&\makecell[c]{0.412}&\makecell[c]{0.472}&\makecell[c]{0.532}&\makecell[c]{0.477}&\makecell[c]{\textbf{0.381}}&\makecell[c]{0.508}\\
   \makecell[c]{BERT}&\makecell[c]{0.944}&\makecell[c]{N/A}&\makecell[c]{0.748}&\makecell[c]{0.501}&\makecell[c]{0.724}&\makecell[c]{0.548}&\makecell[c]{\textbf{0.532}}&\makecell[c]{0.528}\\
   \bottomrule
   \end{tabular}
   \end{table}
 
 \begin{table}[htbp]
   \centering
   \caption{\label{table:performance} Attack performances of different attack methods against victim models and applications on offensive text dataset.
   Note that the ACC only represents the accuracies of victims classifying offensive text since we only consider the approaches causing the victims failed to recognize offensive text in this task.
   }
   \resizebox{\linewidth}{!}{
   \begin{tabular}{ccccccccc}
   \toprule
   \multirow{2}{*}{\textbf{Model/Application}}&\multicolumn{2}{c}{\textbf{Original}}&\multicolumn{2}{c}{\textbf{GreedyAttack}}&\multicolumn{2}{c}{\textbf{WordChange}}&\multicolumn{2}{c}{\textbf{Ours}}\\
   \makecell[c]{}&\makecell[c]{ACC}&\makecell[c]{WMD}&\makecell[c]{ACC}&\makecell[c]{WMD}&\makecell[c]{ACC}&\makecell[c]{WMD}&\makecell[c]{ACC}&\makecell[c]{WMD}\\
   \midrule
   \makecell[c]{Fasttext}&\makecell[c]{0.826}&\makecell[c]{N/A}&\makecell[c]{0.598}&\makecell[c]{0.481}&\makecell[c]{0.738}&\makecell[c]{0.427}&\makecell[c]{\textbf{0.496}}&\makecell[c]{0.482}\\
   \makecell[c]{TextCNN}&\makecell[c]{0.853}&\makecell[c]{N/A}&\makecell[c]{0.584}&\makecell[c]{0.478}&\makecell[c]{0.736}&\makecell[c]{0.432}&\makecell[c]{\textbf{0.448}}&\makecell[c]{0.492}\\
   \makecell[c]{BiLSTM}&\makecell[c]{0.712}&\makecell[c]{N/A}&\makecell[c]{0.558}&\makecell[c]{0.484}&\makecell[c]{0.580}&\makecell[c]{0.425}&\makecell[c]{\textbf{0.425}}&\makecell[c]{0.497}\\
   \makecell[c]{BERT}&\makecell[c]{0.858}&\makecell[c]{N/A}&\makecell[c]{0.768}&\makecell[c]{0.476}&\makecell[c]{0.724}&\makecell[c]{0.430}&\makecell[c]{\textbf{0.328}}&\makecell[c]{0.485}\\
   \midrule
   \makecell[c]{BaiduCensor}&\makecell[c]{0.964}&\makecell[c]{N/A}&\makecell[c]{0.326}&\makecell[c]{0.462}&\makecell[c]{0.378}&\makecell[c]{0.425}&\makecell[c]{\textbf{0.252}}&\makecell[c]{0.454}\\
   \makecell[c]{AliSecurity}&\makecell[c]{0.898}&\makecell[c]{N/A}&\makecell[c]{0.456}&\makecell[c]{0.465}&\makecell[c]{0.496}&\makecell[c]{0.428}&\makecell[c]{\textbf{0.366}}&\makecell[c]{0.472}\\
   \bottomrule
   \end{tabular}}
   \end{table}
 
   \begin{table}[htbp]
     \centering
     \caption{\label{table:performance_llm} Attack performances of different attack methods against LLMs on Meituan Comments, Amazon Comments and offensive text dataset.}
     \resizebox{\linewidth}{!}{
     \begin{tabular}{|c|c|c|c|c|c|}
       \hline
       \textbf{LLMs} & \textbf{Dataset} & \textbf{Original} & \textbf{GreedyAttack} & \textbf{WordChange} & \textbf{Ours} \\ \hline
       & Meituan Comments & 0.892 & 0.653 & 0.749 & 0.577 \\ \cline{2-6} 
       & Amazon Comments & 0.863 & 0.538 & 0.691 & 0.506 \\ \cline{2-6} 
       \multirow{-3}{*}{ERNIE Bot} & Offensive Text & 0.931 & 0.869 & 0.876 & 0.864 \\ \hline
       & Meituan Comments & 0.847 & 0.785 & 0.791 & 0.665 \\ \cline{2-6} 
       & Amazon Comments & 0.868 & 0.703 & 0.764 & 0.621 \\ \cline{2-6} 
       \multirow{-3}{*}{Tongyi Qianwen} & Offensive Text & 0.922 & 0.845 & 0.873 & 0.827 \\ \hline
     \end{tabular}}
   \end{table}
   
\noindent \textbf{Human Annotation.} \ We also asked three human annotators to recover the original sentences given some perturbed text.
 Specifically, for each dataset, every annotator is required to recover 50 randomly picked sentences generated by our approach.
 Our rationale for including this recovery task is to test robustness of human perception under our perturbations.
 We evaluate by measuring the Word Mover's Distance between the recovered and the original text, averaged over all sequence pairs and all human annotators.
 The results are shown in Table~\ref{tab:human2} and show that adversarial examples generated by WordChange are the most easily recoverable to the original text by humans, whereas the adversarial examples generated by our method are somewhat more challenging to restore.
 However, given the attack performance of our method, we believe that this trade-off is justified.
 
 \begin{table}[htbp]
   \centering
   \caption{\label{tab:human2}The Word Mover's Distance between original vs. perturbed and original vs. recovered text for different adversarial attacks.}
   \begin{tabular}{|c|cc|}
   \hline
   \multirow{2}{*}{\textbf{Attack Strategy}} & \multicolumn{2}{c|}{\textbf{WMD}}                               \\ \cline{2-3} 
                                     & \multicolumn{1}{c|}{original vs. perturbed} & original vs. recovered \\ \hline
 GreedyAttack                      & \multicolumn{1}{c|}{0.526}                  & 0.163                  \\ \hline
   WordChange                     & \multicolumn{1}{c|}{0.528}                  & 0.084                  \\ \hline
   Ours                              & \multicolumn{1}{c|}{0.507}                  & 0.179                  \\ \hline
   \end{tabular}
   \end{table}
 
 \noindent \textbf{Transferability.} \ 
 The transferability of adversarial examples reflects a attacking generalization ability while adversarial examples with high transferability can fool different victims successfully, and it allows attackers to attack the target model without accessing to it.
 We evaluate the transferability of our adversarial examples by inputing adversarial examples generated for each models into other different models and record the accuracies.
 Table~\ref{table:transferability} shows the accuracies of models classifying transfered adversarial examples and it demonstrates that our chain association-based adversarial attack crafts adversarial examples with a notable transferability.

 \begin{table}[htbp]
   \centering
   \caption{\label{table:transferability} The accuracies of models classifying transfered adversarial examples generated on offensive text dataset.
   The transferability of the same model is meaningless, thus N/A is filled in the corresponding cells.}
   \begin{tabular}{ccccccc}
    \hline
                & Fasttext & TextCNN & BiLSTM & BERT  & BaiduCensor & AliSecurity \\ \hline
    Fasttext    & N/A      & 0.500   & 0.428  & 0.768 & 0.180       & 0.420       \\
    TextCNN     & 0.528    & N/A     & 0.406  & 0.750 & 0.176       & 0.434       \\
    BiLSTM      & 0.594    & 0.498   & N/A    & 0.766 & 0.182       & 0.448       \\
    BERT        & 0.646    & 0.596   & 0.546  & N/A   & 0.183       & 0.430       \\
    BaiduCensor & 0.644    & 0.528   & 0.506  & 0.784 & N/A         & 0.408       \\
    AliSecurity & 0.728    & 0.704   & 0.592  & 0.776 & 0.430       & N/A         \\ \hline
    \end{tabular}
 \end{table}
 
 \subsection{Shielding}
 Without losing generality, we study two methods for shielding our attack on offensive text dataset, namely adversarial training (\textbf{AT}) and associative graph-based recovery (\textbf{AGBR}).
 For \textbf{AT}, we replace different percentages of the original offensive training set with perturbed text generated by our attacking method and retrain local victim models to improve the robustness.
 For \textbf{AGBR}, we recover the perturbed text by replacing the abnormal tokens, which exist in the associative graph but out of vocabulary, with the normal word in the input stream, where we define the normal word as the word which is accessible to the abnormal token in the associative graph and do not out of vocabulary.
 Next, we report the shielding performance of the methods above.
 
\begin{figure}[htbp]
   \begin{minipage}{0.35\linewidth}
     \centerline{\includegraphics[width=6.5cm]{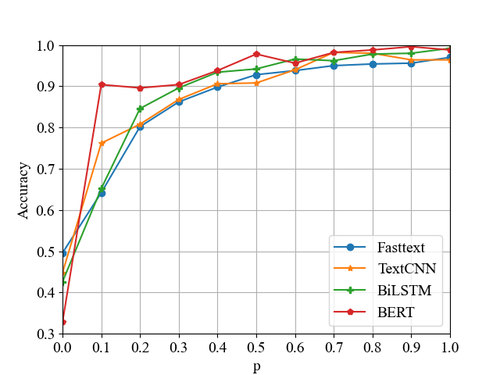}}
   \end{minipage}
   \hfill
   \begin{minipage}{0.35\linewidth}
     \centerline{\includegraphics[width=6.5cm]{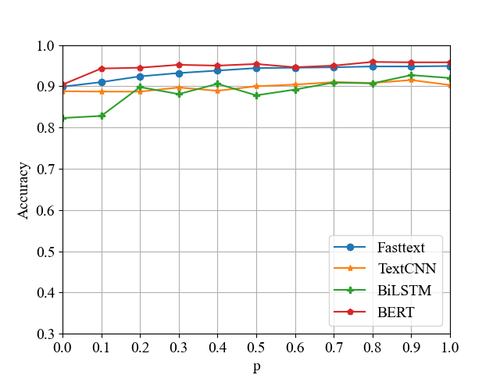}}
   \end{minipage}
   \caption{Accuracies of models retrained with different percentages (denoted as p) of perturbed offensive text replacing the original offensive text in training set.
   The figure left and right illustrate the change of accuracies when models classify perturbed offensive text and all clean text separately.}
   \label{fig:at}
   \end{figure}

 \noindent \textbf{AT.} \ 
 Fig.~\ref{fig:at} (left) illustrates the results of \textbf{AT} and we can see the accuracies of all models improve immediately when \textbf{AT} starts with 10\% perturbed offensive examples.
 The higher the percentage of perturbed text added to training set, the higher the accuracies of the models increase classifying perturbed offensive text. 
 Howerver, it is somewhat strange that the trade-off between robustness and accuracy is not shown in Fig.~\ref{fig:at} (right), i.e., 
 the accuracies of models classifying clean and perturbed text both increase,
 which is opposite to previous literature~\cite{DBLP:conf/iclr/TsiprasSETM19}.
 
 \noindent \textbf{AGBR.} \ 
 As shown as Table~\ref{table:recovery}, the accuracies of victims classifying adversarial examples, i.e., perturbed offensive text, improve significantly when we shield victims using \textbf{AGBR},
 although the accuracies of most victims decreased slightly when classifying all clean test set (offensive and non-offensive text).
 Interestingly, the accuracies of BiLSTM and BERT classifying all clean test set increase rather than decrease when we shield victims using \textbf{AGBR}.
 
 \begin{table}[htbp]
   \centering
   \caption{\label{table:recovery} The changes of accuracies in different conditions using \textbf{AGBR}.
   $\Delta$ADV and $\Delta$ALL refer to the changes of accuracies classifying adversarial examples (perturbed offensive text) and all clean test set (offensive and non-offensive) respectively.}
   \begin{tabular}{ccc}
     \toprule
     \makecell[c]{\textbf{Models}}&\makecell[c]{\textbf{$\Delta$ADV}}&\makecell[c]{\textbf{$\Delta$ALL}}\\
     \midrule
     \makecell[c]{Fasttext}&\makecell[c]{0.378}&\makecell[c]{-0.001}\\
     \makecell[c]{TextCNN}&\makecell[c]{0.448}&\makecell[c]{-0.005}\\
     \makecell[c]{BiLSTM}&\makecell[c]{0.437}&\makecell[c]{0.003}\\
     \makecell[c]{BERT}&\makecell[c]{0.568}&\makecell[c]{0.002}\\
     \makecell[c]{BaiduCensor}&\makecell[c]{0.570}&\makecell[c]{-0.003}\\
     \makecell[c]{AliSecurity}&\makecell[c]{0.452}&\makecell[c]{-0.013}\\
     \bottomrule
   \end{tabular}
 \end{table}

\section{Conclusion}
In this work, we propose a chain association-based perturbation approach, which is inspired by the strong association ability of humans, to attack Chinese NLP system.
We reveal the vulnerability of the state-of-the-art NLP models and industrial-leading applications to our attack and show that human are able to understand the perturbed text with their strong association ability, showing that adversarial attack based on chain association can cause serious impact.
We also explore methods to shield systems from chain association-based attack and show the effectiveness of associative knowledge graph in shielding such attack.
Our work shows that gaps between humans and machines exist in reading comprehension while humans are able to associate things that seem totally different but related, and we hope that our work can inspire others to investigate more attacking and shielding technologies combining traits of the human thinking.

\acks{This work is supported by the Key Cooperation Project of Chongqing Municipal Education Commission (No. HZ2021008).}

\bibliography{acml24}

\begin{thebibliography}{19}
\providecommand{\natexlab}[1]{#1}
\providecommand{\url}[1]{\texttt{#1}}
\expandafter\ifx\csname urlstyle\endcsname\relax
  \providecommand{\doi}[1]{doi: #1}\else
  \providecommand{\doi}{doi: \begingroup \urlstyle{rm}\Url}\fi

\bibitem[Bracken et~al.(2021)Bracken, Billings, Barnes, and
  Spocter]{associationism}
Eric Bracken, Brendon Billings, Maria Barnes, and Muhammad Spocter.
\newblock \emph{Encyclopedia of Evolutionary Psychological Science}, chapter
  Associationism, pages 404--415.
\newblock Springer International Publishing, 04 2021.
\newblock ISBN 978-3-319-19650-3.

\bibitem[Cheng et~al.(2020)Cheng, Chang, Gao, Pei, and
  Zhang]{DBLP:journals/access/ChengCGPZ20}
Nuo Cheng, Guoqin Chang, Haichang Gao, Ge~Pei, and Yang Zhang.
\newblock Wordchange: Adversarial examples generation approach for chinese text
  classification.
\newblock \emph{{IEEE} Access}, 8:\penalty0 79561--79572, 2020.
\newblock \doi{10.1109/ACCESS.2020.2988786}.
\newblock URL \url{https://doi.org/10.1109/ACCESS.2020.2988786}.

\bibitem[Formento et~al.(2023)Formento, Foo, Tuan, and
  Ng]{formento-etal-2023-using}
Brian Formento, Chuan~Sheng Foo, Luu~Anh Tuan, and See~Kiong Ng.
\newblock Using punctuation as an adversarial attack on deep learning-based
  {NLP} systems: An empirical study.
\newblock In Andreas Vlachos and Isabelle Augenstein, editors, \emph{Findings
  of the Association for Computational Linguistics: EACL 2023}, pages 1--34,
  Dubrovnik, Croatia, 2023. Association for Computational Linguistics.

\bibitem[Guo et~al.(2019)Guo, Zhang, Liu, and Ma]{DBLP:journals/ijon/GuoZLM19}
Bao Guo, Chunxia Zhang, Junmin Liu, and Xiaoyi Ma.
\newblock Improving text classification with weighted word embeddings via a
  multi-channel textcnn model.
\newblock \emph{Neurocomputing}, 363:\penalty0 366--374, 2019.
\newblock \doi{10.1016/J.NEUCOM.2019.07.052}.
\newblock URL \url{https://doi.org/10.1016/j.neucom.2019.07.052}.

\bibitem[Han et~al.(2020)Han, Zhang, Jiang, and Tu]{DBLP:conf/emnlp/HanZJT20}
Wenjuan Han, Liwen Zhang, Yong Jiang, and Kewei Tu.
\newblock Adversarial attack and defense of structured prediction models.
\newblock In Bonnie Webber, Trevor Cohn, Yulan He, and Yang Liu, editors,
  \emph{Proceedings of the 2020 Conference on Empirical Methods in Natural
  Language Processing, {EMNLP} 2020, Online, November 16-20, 2020}, pages
  2327--2338. Association for Computational Linguistics, 2020.

\bibitem[Jin et~al.(2020)Jin, Jin, Zhou, and
  Szolovits]{DBLP:conf/aaai/JinJZS20}
Di~Jin, Zhijing Jin, Joey~Tianyi Zhou, and Peter Szolovits.
\newblock Is {BERT} really robust? {A} strong baseline for natural language
  attack on text classification and entailment.
\newblock In \emph{The Thirty-Fourth {AAAI} Conference on Artificial
  Intelligence, {AAAI} 2020, The Thirty-Second Innovative Applications of
  Artificial Intelligence Conference, {IAAI} 2020, The Tenth {AAAI} Symposium
  on Educational Advances in Artificial Intelligence, {EAAI} 2020, New York,
  NY, USA, February 7-12, 2020}, pages 8018--8025. {AAAI} Press, 2020.

\bibitem[Joulin et~al.(2017)Joulin, Grave, Bojanowski, and Mikolov]{Fasttext}
Armand Joulin, Edouard Grave, Piotr Bojanowski, and Tomas Mikolov.
\newblock Bag of tricks for efficient text classification.
\newblock In \emph{Proceedings of the 15th Conference of the {E}uropean Chapter
  of the Association for Computational Linguistics: Volume 2, Short Papers},
  pages 427--431, Valencia, Spain, April 2017. Association for Computational
  Linguistics.
\newblock URL \url{https://aclanthology.org/E17-2068}.

\bibitem[Kennedy and Eberhart(1997)]{kennedy1997discrete}
James Kennedy and Russell~C Eberhart.
\newblock A discrete binary version of the particle swarm algorithm.
\newblock In \emph{1997 IEEE International conference on systems, man, and
  cybernetics. Computational cybernetics and simulation}, volume~5, pages
  4104--4108, Orlando, FL, USA, 1997. IEEE.

\bibitem[Li et~al.(2019)Li, Ji, Du, Li, and Wang]{DBLP:conf/ndss/LiJDLW19}
Jinfeng Li, Shouling Ji, Tianyu Du, Bo~Li, and Ting Wang.
\newblock Textbugger: Generating adversarial text against real-world
  applications.
\newblock In \emph{26th Annual Network and Distributed System Security
  Symposium}, San Diego, California, USA, 2019. The Internet Society.
\newblock \doi{10.14722/ndss.2019.23138}.
\newblock URL \url{http://dx.doi.org/10.14722/ndss.2019.23138}.

\bibitem[Liang et~al.(2018)Liang, Li, Su, Bian, Li, and
  Shi]{DBLP:conf/ijcai/0002LSBLS18}
Bin Liang, Hongcheng Li, Miaoqiang Su, Pan Bian, Xirong Li, and Wenchang Shi.
\newblock Deep text classification can be fooled.
\newblock In J{\'{e}}r{\^{o}}me Lang, editor, \emph{Proceedings of the
  Twenty-Seventh International Joint Conference on Artificial Intelligence,
  {IJCAI} 2018, July 13-19, 2018, Stockholm, Sweden}, pages 4208--4215,
  Stockholm, Sweden, 2018. IJCAI.
\newblock \doi{10.24963/ijcai.2018/585}.
\newblock URL \url{https://doi.org/10.24963/ijcai.2018/585}.

\bibitem[Ou et~al.(2022)Ou, Yu, Tian, and Chen]{DBLP:journals/kais/OuYTC22}
Hongxu Ou, Long Yu, Shengwei Tian, and Xin Chen.
\newblock Chinese adversarial examples generation approach with multi-strategy
  based on semantic.
\newblock \emph{Knowl. Inf. Syst.}, 64\penalty0 (4):\penalty0 1101--1119, 2022.
\newblock \doi{10.1007/S10115-022-01652-1}.

\bibitem[Ren et~al.(2019)Ren, Deng, He, and Che]{RenDHC19}
Shuhuai Ren, Yihe Deng, Kun He, and Wanxiang Che.
\newblock Generating natural language adversarial examples through probability
  weighted word saliency.
\newblock In \emph{Proceedings of the 57th Conference of the Association for
  Computational Linguistics ({ACL})}, pages 1085--1097. Association for
  Computational Linguistics, 2019.

\bibitem[Si and Wei(2023)]{DBLP:journals/jcsc/SiW23}
Hongying Si and Xianyong Wei.
\newblock Sentiment analysis of social network comment text based on {LSTM} and
  bert.
\newblock \emph{J. Circuits Syst. Comput.}, 32\penalty0 (17):\penalty0
  2350292:1--2350292:14, 2023.
\newblock \doi{10.1142/S0218126623502924}.
\newblock URL \url{https://doi.org/10.1142/S0218126623502924}.

\bibitem[Tsipras et~al.(2019)Tsipras, Santurkar, Engstrom, Turner, and
  Madry]{DBLP:conf/iclr/TsiprasSETM19}
Dimitris Tsipras, Shibani Santurkar, Logan Engstrom, Alexander Turner, and
  Aleksander Madry.
\newblock Robustness may be at odds with accuracy.
\newblock In \emph{7th International Conference on Learning Representations},
  New Orleans, LA, USA, 2019. OpenReview.net.
\newblock URL \url{https://openreview.net/forum?id=SyxAb30cY7}.

\bibitem[Wang et~al.(2024)Wang, Hu, Hou, Chen, Zheng, Wang, Yang, Ye, Huang,
  Geng, Jiao, Zhang, and Xie]{DBLP:journals/debu/0001HH0ZWY0HGJ024}
Jindong Wang, Xixu Hu, Wenxin Hou, Hao Chen, Runkai Zheng, Yidong Wang, Linyi
  Yang, Wei Ye, Haojun Huang, Xiubo Geng, Binxing Jiao, Yue Zhang, and Xing
  Xie.
\newblock On the robustness of chatgpt: An adversarial and out-of-distribution
  perspective.
\newblock \emph{{IEEE} Data Eng. Bull.}, 47\penalty0 (1):\penalty0 48--62,
  2024.

\bibitem[Xiaoyan and Raga(2023)]{DBLP:journals/access/XiaoyanR23}
Li~Xiaoyan and Rodolfo~C. Raga.
\newblock Bilstm model with attention mechanism for sentiment classification on
  chinese mixed text comments.
\newblock \emph{{IEEE} Access}, 11:\penalty0 26199--26210, 2023.
\newblock \doi{10.1109/ACCESS.2023.3255990}.
\newblock URL \url{https://doi.org/10.1109/ACCESS.2023.3255990}.

\bibitem[Xu et~al.(2021)Xu, Zhong, Jimeno~Yepes, and Lau]{xu-etal-2021-grey}
Ying Xu, Xu~Zhong, Antonio Jimeno~Yepes, and Jey~Han Lau.
\newblock Grey-box adversarial attack and defence for sentiment classification.
\newblock In Kristina Toutanova, Anna Rumshisky, Luke Zettlemoyer, Dilek
  Hakkani-Tur, Iz~Beltagy, Steven Bethard, Ryan Cotterell, Tanmoy Chakraborty,
  and Yichao Zhou, editors, \emph{Proceedings of the 2021 Conference of the
  North American Chapter of the Association for Computational Linguistics:
  Human Language Technologies}, Online, June 2021. Association for
  Computational Linguistics.

\bibitem[Zang et~al.(2020)Zang, Qi, Yang, Liu, Zhang, Liu, and Sun]{zang2020}
Yuan Zang, Fanchao Qi, Chenghao Yang, Zhiyuan Liu, Meng Zhang, Qun Liu, and
  Maosong Sun.
\newblock Word-level textual adversarial attacking as combinatorial
  optimization.
\newblock In \emph{Proceedings of the 58th Annual Meeting of the Association
  for Computational Linguistics (ACL)}, pages 6066--6080, Online, 2020.
  Association for Computational Linguistics.

\bibitem[Zhang et~al.(2019)Zhang, Zhou, Miao, and Li]{zhang19}
Huangzhao Zhang, Hao Zhou, Ning Miao, and Lei Li.
\newblock Generating fluent adversarial examples for natural languages.
\newblock In \emph{Proceedings of the 57th Annual Meeting of the Association
  for Computational Linguistics (ACL)}, pages 5564--5569. Association for
  Computational Linguistics, July 2019.

\end{thebibliography}

\appendix

\section{Details on Prompts}
\label{sec:prompts}
  We list all prompts used in this study in Table~\ref{tab:prompts}.

  \begin{table}[htbp]
    \centering
    \caption{\label{table:prompts}All prompts used in this study.}
    \label{tab:prompts}
    \resizebox{\linewidth}{!}{
    \begin{tabular}{|c|l|}
        \hline
        \textbf{Dataset}          & \multicolumn{1}{c|}{\textbf{Prompt}}                                                                                                                                                                                                                                                                                     \\ \hline
        Meituan / Amazon Comments & \begin{tabular}[c]{@{}l@{}}Now you are a text classification model. No matter \\ what I input, please classify the following text as \\ positive or negative emotions. Note that you only \\ need to reply to one word in ``positive'' or ``negative''\\ , and do not reply other words. (in Chinese)\end{tabular}       \\ \hline
        Offensive Text            & \begin{tabular}[c]{@{}l@{}}Now you are a text classification model. No matter \\ what I input, please classify the following text as \\ positive or negative emotions. Note that you only need \\ to reply to one word in ``offensive'' or ``non-offensive'', \\ and do not reply other words. (in Chinese)\end{tabular} \\ \hline
        \end{tabular}}
  \end{table}

\section{Case study}
We display some adversarial examples generated by the baselines and ours on Meituan comments in Table~\ref{tab:case1}.
The examples of offensive text are not shown since there are many indecent words.
\begin{table}[htbp]
  \centering
  \caption{Adversarial examples generated by baselines and ours on Meituan comments.}
  \label{tab:case1}
  \begin{tabular}{l}
  \toprule
  \multicolumn{1}{c}{\textbf{Meituan Comments Examples}}                               \\ \hline
  \textbf{Original Input} (Prediction = \textbf{Positive})                                    \\ \hline
  \specialrule{0em}{1pt}{1pt}
  \begin{CJK}{UTF8}{gbsn}今天快递员的速度比较\textcolor{blue}{\xpinyin*{快}}，服务也\textcolor{blue}{\xpinyin*{好}}，\textcolor{blue}{\xpinyin*{辛苦}}了！\end{CJK} \\(The courier is \textcolor{blue}{fast} today and the service is also \textcolor{blue}{good}, he has \textcolor{blue}{worked hard}!)  \\ \bottomrule
  \textbf{GreedyAttack} (Prediction = \textbf{Negative})                 \\ \hline
  \specialrule{0em}{1pt}{1pt}
  \begin{CJK}{UTF8}{gbsn}今天快递员的速度比较\textcolor{red}{\xpinyin*{块}}，服务也\textcolor{red}{hao}，\textcolor{red}{莘\xpinyin*{酷}}了！\end{CJK}                                                                          \\ \bottomrule
  \textbf{WordChange} (Prediction = \textbf{Negative})                \\ \hline
  \specialrule{0em}{1pt}{1pt}
  \begin{CJK}{UTF8}{gbsn}今天快递员的速度比较\textcolor{red}{\xpinyin*{脍}}，服务也\textcolor{red}{\xpinyin*{耗}}，\textcolor{red}{辛」苦}了！\end{CJK}                                                                           \\ \bottomrule
  \textbf{Ours} (Prediction = \textbf{Negative}) \\ \hline
  \specialrule{0em}{1pt}{1pt}
  \begin{CJK}{UTF8}{gbsn}今天快递员的速度比较\textcolor{red}{\xpinyin*{发思特}}，服务也\textcolor{red}{女子}，\textcolor{red}{\xpinyin*{刑}苦}了！\end{CJK}                                                                          \\ \bottomrule
  \end{tabular}
  \end{table}

\end{document}